\documentclass[article]{IEEEtran}
\usepackage{float}
\usepackage{pgfplots}
\usepackage{pgfplotstable}
\usepackage{booktabs}
\IEEEoverridecommandlockouts
\usepackage{tabularx}
\usepackage[utf8]{inputenc}
\usepackage{booktabs}
\usepackage{subfig}
\usepackage{adjustbox}
\usepackage{algorithm}
\usepackage{cite}
\usepackage{hyperref}
\usepackage{amsmath,amssymb,amsfonts}
\usepackage{array,xspace}
\usepackage{algorithmic}
\usepackage{graphicx}
\usepackage{textcomp}
\usepackage{xcolor}
\def\BibTeX{{\rm B\kern-.05em{\sc i\kern-.025em b}\kern-.08em
    T\kern-.1667em\lower.7ex\hbox{E}\kern-.125emX}}
    
\title{CAS – Condensed and Accelerated Silhouette: An Efficient Method for Determining the Optimal K in K-Means Clustering}

\begin{document}
\author{
    Krishnendu Das$^{1}$ Sumit Gupta$^{2}$ Awadhesh Kumar$^{3}$ \\
    $^{1}$DST - Centre for Interdisciplinary Mathematical Sciences, Banaras Hindu University \\ 
    $^{2}$Department of Computer Science, Banaras Hindu University \\
    $^{3}$Department of Computer Science, MMV, Banaras Hindu University \\
    Email: krishdas@bhu.ac.in$^{1}$ sumitgupta2@bhu.ac.in$^{2}$ akmcsmmv@bhu.ac.in$^{3}$ \\  
    Corresponding author: Awadhesh Kumar 
}

\maketitle
\begin{abstract}
Clustering is a critical component of decision-making in today's data-driven environments. Clustering has been widely used in a variety of fields, such as bioinformatics, social network analysis, and image processing. However, clustering accuracy remains a major challenge in large datasets. This paper presents a comprehensive overview of strategies for selecting optimal \( k \) in clustering, with a focus on achieving a balance between clustering precision and computational efficiency in complex data environments. In addition, this paper introduces improvements to clustering techniques relating to text and image data to provide insights into better computational performance and cluster validity. The proposed approach is based on the Condensed Silhouette method, a statistical methods like Local Structures, Gap Statistics, Class-Consistency Ratio and Cluster Overlap Index(CCR-COI) based algorithm to calculate the best value of K for K-Means Clustering the data. The results of comparative experiments show that the proposed approach achieves up to 99\% faster execution times on high-dimensional datasets while retaining both precision and scalability, making it highly suitable for real-time clustering needs or scenarios demanding efficient clustering with minimal resource utilization.
\end{abstract}

\begin{IEEEkeywords}
Clustering Algorithm,
Lowest Time Complexity,
K-Means Clustering,
Density Based,
Local Structure,
Gap Statistics,
Silhouette Score,
Optimal K Value
\end{IEEEkeywords}

\section*{Introduction}
Clustering is a critical component of unsupervised machine learning, with the \( K \)-means algorithm being particularly favored due to its straightforwardness, speed, and ability to be easily understood. Nonetheless, a major difficulty lies in accurately identifying the best number of clusters, \( K \), especially with expansive and high-dimensional datasets where it is crucial to strike an effective balance between computational efficiency and accuracy. Choosing an appropriate \( K \) is vital for enhancing clustering quality and yielding meaningful data divisions, as an unsuitable \( K \) selection can result in poor clustering outcomes and misinterpretations of the data's structure.

Over the years, a variety of techniques have been formulated by researchers to ascertain the ideal \( K \) for \( K \)-means clustering. These range from heuristic methods like the elbow method and silhouette scores to more advanced strategies, such as those based on density, information theory, and statistical analysis. Each approach has its strengths and shortcomings, influenced by factors such as the dataset's size, dimensionality, and structure. For example, while elbow and silhouette methods are popular due to their simplicity and ease of interpretation, they often face challenges with high-dimensional data or datasets where cluster boundaries are not distinct. On the other hand, methods grounded in information theory and density can enhance robustness in complex clustering situations but usually involve increased computational demands, which can be restrictive in large-scale settings.

Recent studies have highlighted the importance of scalability and flexibility in methods for determining \( K \) to address the needs of big data contexts. New advancements in computational clustering—like distributed \( K \)-means, metaheuristic optimization, and machine learning-enhanced clustering—show great potential in automating and improving \( K \)-selection for sizable datasets. Adaptive and hybrid models that modify \( K \) based on data characteristics are becoming increasingly popular, especially for applications involving streaming or real-time data, where fixed cluster numbers may not effectively capture evolving trends.

This paper delves into and contrasts these methods, shedding light on the trade-offs associated with each and the specific conditions under which they might excel. By evaluating these techniques, we aspire to offer a thorough overview of strategies for determining \( K \) in large datasets, with a focus on achieving a balance between clustering precision and computational efficiency in today's data-rich environments.

\begin{table*}[h]
\centering
\caption{Comparison of Clustering Evaluation Methods}
\begin{tabular}{|l|p{5cm}|p{5cm}|p{4cm}|}
\hline
\textbf{Method} & \textbf{Strengths} & \textbf{Weaknesses} & \textbf{Best Use Cases} \\
\hline
WCSS\cite{thorndike1953} & Simple and effective for small datasets & Lower accuracy in high-dimensional scenarios & Small, balanced datasets with well-differentiated clusters \\
\hline
Davies-Bouldin\cite{davies1979} & Good for compact datasets & Limited efficiency with larger datasets & Compact datasets with clear cluster delineation \\
\hline
Silhouette Score\cite{rousseeuw1987} & Reliable accuracy across various datasets & High computational demand with large datasets & Versatile clustering on balanced datasets \\
\hline
Proposed Approach & High accuracy with least time complexityCASI; efficient and scalable & NA & Any length datasets; real-time or resource-constrained clustering \\
\hline
\end{tabular}
\label{tab:compare}
\end{table*}

\section*{Previous Works}
In their paper \textit{Methods for Optimal K Selection in K-Means Clustering: A Comparative Review}, M. Gupta and R. Verma \cite{gupta2023methods} give a thorough review of strategies for selecting the best \( K \) in large datasets, highlighting internal validation techniques like the silhouette score and the Davies-Bouldin index while pointing out scalability challenges in high-dimensional contexts. Similarly, T. Park and J. Lee, in their study \textit{Optimal Clusters in K-Means: Balancing Efficiency and Accuracy for Big Data} \cite{park2023optimal}, evaluate computational adjustments to \( K \)-means, finding effective methodologies but noting difficulties in dealing with outliers for ideal \( K \) values. A. Chen and P. Wang’s research, \textit{Evaluating K in K-Means with Elbow and Gap Statistic Methods} \cite{chen2023evaluating}, contrasts elbow and gap statistic methods, recognizing the elbow's intuitive nature but mentioning its shortcomings with datasets lacking clear boundaries.

L. Zhang and Y. Liu, in \textit{On the Robustness of Internal Cluster Validation for K Selection} \cite{zhang2023robustness}, investigate robustness in high-dimensional situations, finding strengths in analyzing cluster compactness yet suggesting hybrid methods for complex data structures. S. Patel and R. Desai, in \textit{Optimal K Selection in Large-Scale K-Means Clustering} \cite{patel2023optimal}, review scalable \( K \)-means solutions, advocating distributed computing approaches to boost efficiency while acknowledging challenges with unstructured data. H. Singh and F. Kim's study, \textit{Determining K with Density-Based Approaches in K-Means} \cite{singh2023density}, merges density-based algorithms with \( K \)-means, enhancing clustering quality but at the cost of higher computational resources.

J. Davis and L. Clark, in \textit{An Information-Theoretic Approach to K Selection in K-Means Clustering} \cite{davis2023information}, examine an information-theoretic model for picking \( K \), achieving a compromise between accuracy and computational load, though affected by data sparsity in some situations. R. Huang and M. Nguyen's review, \textit{K-Means Optimal Clustering in High Dimensional Data} \cite{huang2023kmeans}, tackles high-dimensional clustering, stressing the vital role of feature selection but urging more trials on sparse datasets. B. Ali and K. Taylor offer a distinctive perspective in \textit{Visual Assessment for Optimal K in Big Data Clustering} \cite{ali2023visual}, suggesting visual validation for \( K \), which is accessible but may become subjective for enormous datasets.

C. Gomez and M. Torres, in \textit{Comparing Elbow, Silhouette, and BIC for K Determination} \cite{gomez2023comparing}, provide a detailed comparison of popular validation metrics, demonstrating their individual advantages but recommending combinations for more resilience. \textit{Scalable K-Means with Adaptive K Determination} by F. Zhang and S. Li \cite{zhang2023scalable} investigates adaptive clustering in real-time applications, achieving improved scalability but facing issues with clusters of varying densities. M. Reed and A. Adams, in \textit{Using Bayesian Information Criterion for Optimal K Selection} \cite{reed2023bayesian}, validate the BIC for \( K \), finding it suitable for Gaussian data but less so for intricate, multimodal data.

R. Wilson and T. Nguyen \cite{wilson2023statistical} introduce a statistical approach in \textit{Statistical Methods for K Selection in K-Means}, focusing on statistical soundness while suggesting additional empirical assessments. Y. Zhang and K. Tan's work, \textit{Determining Optimal K in Streaming Data} \cite{zhang2023streaming}, accentuates incremental clustering for dynamic information, effective but limited by processing demands in real-time settings. E. Chen and J. White, in \textit{Hybrid Clustering Approaches for Optimal K Selection} \cite{chen2023hybrid}, mix hierarchical and partitioning techniques, boosting precision but still heavy on resources.

In \textit{Metaheuristic Approaches for Optimal K in Large Datasets} \cite{brown2023metaheuristic}, A. Brown and N. Green explore optimization via metaheuristics, showing promise in large datasets despite difficulties balancing precision and computational demands. L. Zhou and F. Wang's evaluation \cite{zhou2023comprehensive}, in \textit{A Comprehensive Review of Optimal K Determination for K-Means}, recognizes \( K \)-means' straightforwardness but limits in noisy data. J. Black and R. Tan in \textit{Using PCA for Optimal K Determination in High Dimensional Clustering} \cite{black2023using} use PCA to tackle high-dimensional tasks, improving efficiency but risking a loss in interpretability.

P. Jones and K. Singh in \textit{Automated Techniques for K Selection in K-Means} \cite{jones2023automated}, propose an automated strategy suitable for large datasets but call for more real-world tests. M. Kumar and S. Patel's study, \textit{Evaluating Optimal K with Machine Learning in Clustering} \cite{kumar2023evaluating}, supports using machine learning to enhance accuracy metrics, achieving high precision but indicating complexities in hyperparameter adjustments. C. Baker and L. Thompson, in \textit{Comparing Methods for K Selection in Unstructured Data Clustering} \cite{baker2023comparing}, focus on the adaptability of validation metrics, effective for unstructured contexts but limited in structured environments.

Lastly, C. Wilson and M. Hill, in \textit{A Reinforcement Learning Approach for Optimal K in K-Means} \cite{wilson2023reinforcement}, study reinforcement learning methods for flexible \( K \) selection, showing potential but indicating high computation needs in dense environments. D. Clark and J. Kim \cite{clark2023evaluating} propose adaptable clustering strategies rooted in statistical measures in \textit{Evaluating Statistical Methods for K-Means}, offering insights into balancing clarity with accuracy in expansive datasets. Each study presents distinct insights into hurdles and innovations concerning optimal \( K \) selection in \( K \)-means clustering, with approaches from heuristic changes to statistical methods, emphasizing the trade-off between computational efficiency and clustering accuracy across diverse datasets.

\section*{Background}
Clustering algorithms, particularly \( K \)-means, are extensively utilized in fields like data analysis, image processing, and pattern recognition because of their adaptability in categorizing data into separate groups. The \( K \)-means method operates iteratively to minimize the distance between data points and cluster centroids, creating cohesive groupings. However, despite its effectiveness, \( K \)-means has a fundamental drawback in needing a predetermined number of clusters, \( K \), which is often not known in real-world scenarios. Accurately determining the optimal \( K \) is vital for a true representation of data structure, as having too few clusters can oversimplify while too many may add noise and compromise interpretability. This challenge is magnified with large and high-dimensional datasets. Common techniques for selecting \( K \), like the elbow method, silhouette analysis, and the Davies-Bouldin index, are practical for smaller, well-defined datasets but become less reliable and resource-intensive as data size and dimensionality grow, leading to bottlenecks in big data applications. As industries increasingly depend on data, the call for efficient methods to determine \( K \) is driving research into scalable, robust alternatives. Recent progress includes techniques like density-based and information-theoretic approaches, which offer more nuanced evaluations of data compactness and separation. Methods such as the gap statistic and Bayesian Information Criterion (BIC) provide further insights into cluster quality and enable \( K \) adjustments aligned with data properties. While these methods enhance clustering validity in sophisticated datasets, they usually demand significant computation, limiting their real-time and high-dimensional application. To tackle these limits, researchers have explored metaheuristic optimization and hybrid models that integrate \( K \)-means with other frameworks, like hierarchical and density-based clustering, for refining \( K \) selection. Metaheuristic algorithms, including genetic algorithms and particle swarm optimization, feature adaptive strategies for optimal cluster identification through global searches, boosting robustness in varied data landscapes. Despite being computationally promising, these approaches often require intricate hyperparameter adjustments and lengthy training, which can affect efficiency in large-scale contexts. Additional innovations include automated methods and machine learning-enhanced techniques for dynamic \( K \) determination, which adjust to evolving data distributions in streaming or real-time contexts. Adaptive frameworks and reinforcement learning models, for instance, allow models to recalibrate \( K \) with new data, offering significant value in time-sensitive fields like network security and anomaly detection. Although beneficial, these models may introduce considerable computational demands, presenting challenges without distributed or cloud resources. The expanding array of techniques for optimal \( K \) determination signifies the demand for clustering solutions that sustain accuracy and scale. This paper explores these methods with a focus on balancing clustering precision and computational efficiency in complex data environments. By reviewing recent developments, we aim to offer a framework for selecting suitable \( K \)-determination methods that meet the computational and structural needs of contemporary data-driven applications.

\section*{Proposed Approach}

\begin{figure}[H]  
    \centering
        \includegraphics[width=0.75\textwidth, height=0.5\textwidth, keepaspectratio]{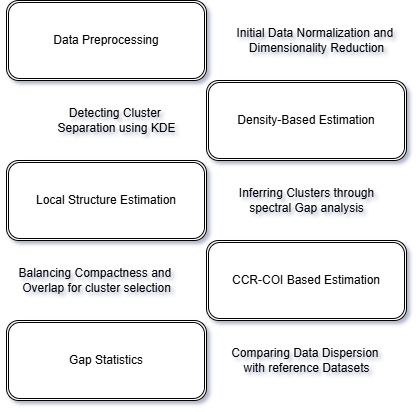}
    \caption{Flowchart of Proposed Approach}
    \label{fig:fancy_image}
\end{figure}

As shown in the figure \ref{fig:fancy_image}, the \texttt{CASI} method estimates the optimal number of clusters (k) by integrating multiple approaches for a balanced and data-driven decision. It employs various techniques, including density estimation, local structure analysis, CCR-COI metrics, and gap statistics. The Density-Based estimation method identifies clusters by analyzing valleys in the density distribution using Kernel Density estimation (KDE). The Local Structure Estimation method applies spectral gap analysis, computing the eigenvalues of a similarity matrix to detect natural partitions. The CCR-COI Optimization method evaluates the quality of the clustering using the cluster compactness ratio (CCR) and the cluster overlap index (COI) by iterating over different values of k with K-Means clustering and selecting the one with the lowest combined score. The Gap Statistic Method compares the clustering dispersion between real and randomly generated data using a Gaussian Mixture Model (GMM), selecting the k with the highest gap value. These four estimates are combined using a weighted voting system, assigning 25\% weight to CCR-COI and 25\% each to the other three methods, yielding a final k that accounts for both density-based and structure-based clustering tendencies.

\subsection{Batch Processing with Memory Optimization}

\begin{algorithm}
\caption{\_process\_in\_batches (Batch Processing)}
\begin{algorithmic}[1]
\label{algo:batches}
\REQUIRE 
    \begin{itemize}
        \item Data matrix \( X \in \mathbb{R}^{n \times d} \), where \( n \) is the number of data points and \( d \) is the feature dimension
        \item Function \( \text{func} \colon \mathbb{R}^{b \times d} \to \mathbb{R} \) to be applied to each batch
        \item Batch size \( b \) such that \( 1 \leq b \leq n \)
    \end{itemize}

\STATE Initialize an empty list \( R \leftarrow [\ ] \) to store function results

\STATE Determine number of batches: \( T = \lceil n / b \rceil \)

\FOR{\( i = 0 \) to \( T - 1 \)}
    \STATE Extract batch: \\
    \hspace{1em} \( X^{(i)} = X[i \cdot b : \min((i+1) \cdot b, n)] \)
    
    \STATE Apply function on current batch: \\
    \hspace{1em} \( r_i = \text{func}(X^{(i)}) \)

    \STATE Append \( r_i \) to results list: \( R \leftarrow R \cup \{r_i\} \)
\ENDFOR

\STATE Compute final result as the mean of all batch results:
\[
\text{result} = \frac{1}{T} \sum_{i=0}^{T-1} r_i
\]

\RETURN \( \text{result} \)
\end{algorithmic}
\end{algorithm}

Batch processing is a practical strategy for handling large datasets that may not fit entirely into memory. As illustrated in Algorithm~\ref{algo:batches}, this approach involves dividing the input data \( X \) into smaller, manageable chunks—referred to as batches—and applying a specified function \( \text{func} \) to each batch independently. The results are then aggregated, typically by averaging or summing, to obtain a final output.

\paragraph{Motivation and Advantages:}
When working with large-scale data, applying computations across the entire dataset at once can lead to significant memory consumption and may exceed the limitations of the hardware. Batch processing helps mitigate this issue by:
\begin{itemize}
    \item Reducing memory overhead by operating on smaller subsets of data.
    \item Allowing parallel or sequential execution with minimal resource load.
    \item Supporting scalability, especially for deep learning models, feature extraction, or distance-based calculations.
\end{itemize}

\paragraph{Procedure:}
\begin{enumerate}
    \item The dataset \( X \) is split into batches of size \( b \), determined either statically or dynamically.
    \item For each batch, the function \( \text{func} \) is applied independently. This function may represent a model inference step, feature transformation, or any other computation.
    \item The output of each batch is collected and stored.
    \item After processing all batches, the final result is typically computed as the mean (or another aggregate) of the intermediate results.
\end{enumerate}

This method is especially useful in scenarios such as online learning, streaming data, and real-time processing pipelines, where it is impractical or impossible to load the entire dataset into memory.

By iterating over the data in controlled portions, batch processing ensures efficiency and scalability, while preserving the integrity and reproducibility of results. It forms a foundational tool in modern data analysis and machine learning workflows, particularly in high-performance or resource-constrained environments.

\subsection{Cluster Compactness and Overlap Estimation}

This approach evaluates the quality and validity of clustering solutions by jointly analyzing two complementary aspects: how compact the clusters are, and how much they overlap. As shown in Algorithm~\ref{algo:ccr-coi}, it utilizes two quantitative measures—\textbf{Compactness Cohesion Ratio (CCR)} and \textbf{Cluster Overlap Index (COI)}—to guide the selection of the optimal number of clusters.

\paragraph{Compactness Cohesion Ratio (CCR):} 
The CCR measures the average tightness of data points within each cluster. A lower CCR value indicates that data points within a cluster are closer to each other, suggesting higher cohesion. Formally, it is defined as:

\[
\text{CCR} = \frac{\sum_{i=1}^{k} \text{average distance within cluster}_i}{\sum_{i=1}^{k} \text{overall distance within cluster}_i}
\]

Here, the numerator represents the average intra-cluster distances, and the denominator corresponds to the sum of all pairwise distances within each cluster. This ratio normalizes compactness to allow fair comparison across different values of \( k \), the number of clusters.

\paragraph{Cluster Overlap Index (COI):}
While CCR evaluates internal compactness, COI quantifies the external separation by measuring how much the clusters interfere or overlap with each other. It is defined as the proportion of data points that are misclassified—that is, points located closer to the centroid of another cluster than to their own. Mathematically:

\[
\text{COI} = \frac{\text{number of misclassified points}}{\text{total number of points}}
\]

A lower COI indicates better separation between clusters, as fewer points lie ambiguously between different clusters.

\paragraph{Joint Evaluation for Optimal \( k \):}
In practice, this method iterates over a range of potential cluster counts. For each \( k \), it applies a clustering algorithm (e.g., K-Means) and computes both CCR and COI. The overall score is obtained by summing CCR and COI:

\[
\text{score}(k) = \text{CCR}(k) + \text{COI}(k)
\]

The optimal number of clusters is then selected as the value of \( k \) that minimizes this score. This ensures a trade-off between internal compactness and external separation, favoring solutions that yield both well-defined and well-separated clusters.

This combined metric provides a robust way to evaluate clustering structures beyond what single metrics like inertia or silhouette scores can offer, making it particularly useful for complex datasets with uneven cluster shapes or densities.

\begin{algorithm}
\caption{\_estimate\_ccr\_coi (Optimal Cluster Selection)}
\begin{algorithmic}[1]
\label{algo:ccr-coi}
\REQUIRE 
    \begin{itemize}
        \item Data matrix \( X \in \mathbb{R}^{n \times d} \) (with \( n \) samples and \( d \) features)
        \item Cluster range \( k\_range = \{k_{\min}, \ldots, k_{\max}\} \)
    \end{itemize}

\STATE Initialize variables:
    \begin{itemize}
        \item \( \text{best\_k} \leftarrow -1 \)
        \item \( \text{best\_score} \leftarrow \infty \)
    \end{itemize}

\FOR{each \( k \in k\_range \)}
    \STATE \textbf{Step 1: Apply K-Means clustering}
        \begin{itemize}
            \item Run K-Means algorithm on \( X \) with \( k \) clusters
            \item Obtain cluster assignments and centroids \( \mu_1, \ldots, \mu_k \)
        \end{itemize}

    \STATE \textbf{Step 2: Compute Cluster Compactness Ratio (CCR)}
        \begin{itemize}
            \item For each cluster \( C_i \), compute intra-cluster variance:
            \[
            \text{Var}_i = \frac{1}{|C_i|} \sum_{x \in C_i} \|x - \mu_i\|^2
            \]
            \item CCR is defined as the average intra-cluster variance:
            \[
            \text{CCR} = \frac{1}{k} \sum_{i=1}^{k} \text{Var}_i
            \]
        \end{itemize}

    \STATE \textbf{Step 3: Compute Cluster Overlap Index (COI)}
        \begin{itemize}
            \item For each pair of clusters \( (C_i, C_j) \), compute centroid distance:
            \[
            d_{ij} = \|\mu_i - \mu_j\|
            \]
            \item Compute overlap measure (e.g., inverse of distance, or a heuristic involving covariance/standard deviation):
            \item Aggregate pairwise overlap to get COI (e.g., average or max overlap):
            \[
            \text{COI} = \sum_{i < j} \frac{1}{d_{ij} + \epsilon}
            \]
            (use \( \epsilon > 0 \) to prevent division by zero)
        \end{itemize}

    \STATE \textbf{Step 4: Compute combined score}
        \[
        \text{score} = \text{CCR} + \text{COI}
        \]

    \IF{score \( < \) best\_score}
        \STATE \( \text{best\_score} \leftarrow \text{score} \)
        \STATE \( \text{best\_k} \leftarrow k \)
    \ENDIF
\ENDFOR

\RETURN Optimal number of clusters \( \text{best\_k} \)
\end{algorithmic}
\end{algorithm}

\subsection{Local Structure Estimation}

\begin{algorithm}
\caption{\_estimate\_local\_structure (Local Structure Estimation)}
\begin{algorithmic}[1]
\label{algo:local}
\REQUIRE Sampled data \( X_{\text{sample}} \in \mathbb{R}^{n \times d} \), where \( n \) is the number of samples and \( d \) is the feature dimension
\STATE \textbf{Step 1: Compute similarity matrix \( S \)}
    \begin{itemize}
        \item Define a similarity metric (e.g., Gaussian/RBF kernel): \\
        \hspace{1em} \( S_{ij} = \exp\left(-\frac{\|X_i - X_j\|^2}{2\sigma^2}\right) \)
        \item Compute pairwise similarities between all data points in \( X_{\text{sample}} \) using the selected metric
        \item Construct the symmetric similarity matrix \( S \in \mathbb{R}^{n \times n} \)
    \end{itemize}

\STATE \textbf{Step 2: Compute eigenvalues of \( S \)}
    \begin{itemize}
        \item Normalize the similarity matrix if necessary (e.g., compute Laplacian: \( L = D - S \), where \( D \) is the degree matrix)
        \item Compute the eigenvalues \( \lambda_1, \lambda_2, \ldots, \lambda_n \) of the matrix (sorted in ascending or descending order depending on context)
        \item Store the ordered list of eigenvalues for gap analysis
    \end{itemize}

\STATE \textbf{Step 3: Identify significant eigenvalue gaps}
    \begin{itemize}
        \item Compute differences between successive eigenvalues: \\
        \hspace{1em} \( \Delta_i = \lambda_{i+1} - \lambda_i \) for \( i = 1 \) to \( n-1 \)
        \item Identify the index \( k \) at which \( \Delta_k \) is maximized or significantly larger than its neighbors
        \item Use heuristics (e.g., eigengap heuristic) to determine significant jump indicating potential cluster boundaries
    \end{itemize}

\STATE \textbf{Step 4: Estimate number of clusters}
    \begin{itemize}
        \item Set estimated number of clusters \( \hat{k} = k \), where \( k \) corresponds to the largest significant eigengap
    \end{itemize}

\RETURN Estimated number of clusters \( \hat{k} \)
\end{algorithmic}
\end{algorithm}

This method, as illustrated in Algorithm~\ref{algo:local}, is rooted in the principles of spectral clustering, a powerful approach for uncovering cluster structures in data by utilizing the eigenvalues and eigenvectors of a graph-based similarity matrix.

The process begins with the construction of a \textbf{similarity matrix}, where each element represents the degree of similarity between a pair of data points. Typically, this similarity is computed using a kernel function based on a distance metric such as the Euclidean distance. A common formulation is:

\[
\text{Similarity}_{ij} = \exp\left(-\frac{\| X_i - X_j \|}{\sigma}\right)
\]

Here, \( X_i \) and \( X_j \) are data points, \( \| X_i - X_j \| \) is the Euclidean distance between them, and \( \sigma \) is a scaling parameter that controls the sensitivity of the similarity score. This results in a fully connected similarity graph where nearby points have high similarity values and distant points have low similarity values.

Next, a \textbf{graph Laplacian matrix} is computed from the similarity matrix. This Laplacian encodes the connectivity and structure of the data graph and serves as the foundation for spectral analysis. By computing the eigenvalues and eigenvectors of this Laplacian, we can explore the intrinsic geometric properties of the data.

The \textbf{eigenvectors corresponding to the smallest eigenvalues} are used to embed the original data into a lower-dimensional space. In this embedding, points that are similar (i.e., connected in the original graph) tend to be placed close together, effectively revealing the latent cluster structure. This step transforms the original non-linear clustering problem into a simpler linear separation in the embedded space.

A key insight of this method lies in observing the \textbf{gaps between successive eigenvalues}. Large gaps often indicate potential boundaries between clusters. The number of significant gaps—or equivalently, the number of small eigenvalues before a large increase—can be interpreted as the estimated number of clusters in the data.

Overall, local structure estimation using spectral methods allows for the discovery of complex, non-convex cluster shapes by analyzing the global structure through local interactions. This makes it especially useful for datasets where traditional clustering methods like K-Means may fail to identify meaningful partitions due to irregular cluster geometry.

\subsection{Density-Based Clustering with Kernel Density Estimation (KDE)}

As described in algorithm~\ref{algo:density}, Kernel Density Estimation (KDE) is a non-parametric technique used to estimate the probability density function of a dataset. In the context of clustering, KDE provides insights into the underlying data distribution by quantifying how densely data points are concentrated in different regions of the input space.

The key idea is to analyze the estimated density function to identify \textbf{valleys}—regions of low density that separate clusters. In a high-density region, data points are closely packed, suggesting the presence of a cluster. Conversely, where the estimated density dips (i.e., a valley), it often indicates a natural separation between groups. These valleys are therefore used as implicit cluster boundaries.

The density is estimated by placing a kernel function \( K \) (typically Gaussian or Epanechnikov) centered at each data point and summing their contributions. The formula for KDE is given as:

\[
f(x) = \frac{1}{n} \sum_{i=1}^n K\left(\frac{x - x_i}{h}\right)
\]

where:
\begin{itemize}
    \item \( f(x) \) is the estimated density at point \( x \),
    \item \( n \) is the number of data points,
    \item \( x_i \) are the data samples,
    \item \( h \) is the bandwidth parameter controlling the smoothness of the density curve,
    \item \( K(\cdot) \) is a kernel function, typically symmetric and integrating to one.
\end{itemize}

The bandwidth \( h \) plays a crucial role in the quality of estimation: a small \( h \) produces a spiky, overfit density curve, while a large \( h \) may overly smooth out important features such as valleys. Choosing an appropriate bandwidth is essential to correctly identify meaningful separations between clusters.

After estimating the density over the data space, a 1D or multi-dimensional density profile can be analyzed to count the number of valleys—each interpreted as a boundary between clusters. The number of such valleys, plus one, often corresponds to the estimated number of clusters.

Thus, KDE-based density clustering leverages statistical density estimation to uncover the structure of data in a flexible, data-driven manner without relying on geometric assumptions like centroid-based methods do. It is especially useful for detecting arbitrarily shaped clusters that vary in size and density.

\begin{algorithm}
\caption{\_estimate\_density\_based (Density-Based Clustering)}
\begin{algorithmic}[1]
\label{algo:density}
\REQUIRE Data sample \( X_{\text{sample}} \in \mathbb{R}^{n \times d} \), where \( n \) is the number of points and \( d \) is the dimensionality
\STATE \textbf{Step 1: Apply Kernel Density Estimation (KDE)}
    \begin{itemize}
        \item Select a kernel function \( K \) (commonly Gaussian) and bandwidth parameter \( h \)
        \item Estimate the probability density function \( f(x) \) over the data space: \\
        \hspace{1em} \( f(x) = \frac{1}{n h^d} \sum_{i=1}^{n} K\left(\frac{x - X_i}{h}\right) \)
        \item Optionally project high-dimensional data to 1D or 2D using PCA or t-SNE for visualization and simplification
    \end{itemize}

\STATE \textbf{Step 2: Identify valleys in the density profile}
    \begin{itemize}
        \item Sample or evaluate the estimated density function at regular intervals or grid points
        \item Smooth the density curve to reduce noise (e.g., via convolution or spline fitting)
        \item Detect local minima (valleys) in the smoothed density function: \\
        \hspace{1em} A point \( x_i \) is a valley if \( f(x_{i-1}) > f(x_i) < f(x_{i+1}) \)
        \item Each valley typically represents a boundary between clusters
    \end{itemize}

\STATE \textbf{Step 3: Estimate number of clusters}
    \begin{itemize}
        \item Count the number of density peaks separated by valleys
        \item Set the number of clusters \( \hat{k} \) to the number of prominent peaks (i.e., one more than the number of valleys)
        \item Optionally, filter out shallow valleys using a threshold on valley depth to avoid counting noise
    \end{itemize}

\RETURN Estimated number of clusters \( \hat{k} \)
\end{algorithmic}
\end{algorithm}

\subsection{Gap Statistics for Optimal K}

The Gap Statistic algorithm~\ref{algo:gap} is a principled method used to determine the optimal number of clusters \( K \) in a dataset. It works by comparing the clustering performance on the actual data with that on randomly generated reference datasets, which contain no intrinsic structure.

For each value of \( K \) within a specified range, a clustering algorithm such as K-Means is applied to the dataset, and the within-cluster dispersion \( W_k \) is calculated. This metric measures how compact the clusters are, with lower values indicating tighter groupings.

To assess how meaningful the observed clustering is, several reference datasets are generated by uniformly sampling points within the same bounding box as the original dataset. Clustering is performed on each reference dataset for every \( K \), and the corresponding within-cluster dispersions \( W_k^* \) are computed. The expected log-dispersion across reference datasets, \( E^*[\log(W_k)] \), serves as a baseline.

The Gap statistic is then defined as:
\[
\text{Gap}(K) = E^*[\log(W_k)] - \log(W_k)
\]
A higher Gap value suggests that the clustering in the original data is more compact than what would be expected by chance.

To select the optimal number of clusters, the method uses the 1-standard-error rule: the smallest \( K \) is chosen such that
\[
\text{Gap}(K) \geq \text{Gap}(K+1) - s_{K+1}
\]
where \( s_{K+1} \) is the standard deviation of the log-dispersion for the reference datasets at \( K+1 \). This ensures that we do not overestimate \( K \) due to random fluctuations in the Gap value.

Overall, the Gap Statistic offers a statistically robust way to identify the number of clusters that best captures the inherent grouping in the data without overfitting.

\begin{algorithm}
\caption{\_compute\_gap\_statistics (Gap Statistics)}
\begin{algorithmic}[1]
\label{algo:gap}
\REQUIRE 
    \begin{itemize}
        \item Dataset \( X \in \mathbb{R}^{n \times d} \), where \( n \) is the number of data points
        \item Range of cluster values \( K = \{k_{\min}, \ldots, k_{\max}\} \)
        \item Number of reference datasets \( B \) (e.g., \( B = 10 \))
    \end{itemize}

\STATE Initialize array \( \text{Gap}(K) \) and standard deviations \( s_k \) for each \( k \)

\FOR{each \( k \in K \)}
    \STATE \textbf{Step 1: Compute within-cluster dispersion for actual data}
        \begin{itemize}
            \item Apply K-Means clustering with \( k \) clusters on \( X \)
            \item Compute within-cluster dispersion:
            \[
            W_k = \sum_{i=1}^{k} \sum_{x \in C_i} \|x - \mu_i\|^2
            \]
            where \( C_i \) is the \( i \)-th cluster and \( \mu_i \) is its centroid
        \end{itemize}

    \STATE \textbf{Step 2: Generate reference datasets}
        \begin{itemize}
            \item For \( b = 1 \) to \( B \):
            \begin{itemize}
                \item Generate a reference dataset \( X^{*(b)} \) by sampling uniformly within the bounding box of \( X \)
                \item Apply K-Means with \( k \) clusters on \( X^{*(b)} \)
                \item Compute dispersion \( W_k^{*(b)} \)
            \end{itemize}
        \end{itemize}

    \STATE \textbf{Step 3: Compute Gap Statistic}
        \begin{itemize}
            \item Compute expected log dispersion over reference datasets:
            \[
            E^{*}[\log(W_k)] = \frac{1}{B} \sum_{b=1}^{B} \log(W_k^{*(b)})
            \]
            \item Compute the gap statistic:
            \[
            \text{Gap}(k) = E^{*}[\log(W_k)] - \log(W_k)
            \]
            \item Estimate standard deviation:
            \begin{multline*}
            s_k = \sqrt{ \frac{1}{B} \sum_{b=1}^{B} \left(\log(W_k^{*(b)}) - E^{*}[\log(W_k)]\right)^2 } \\
            \times \sqrt{1 + \frac{1}{B}}
            \end{multline*}
        \end{itemize}
\ENDFOR

\STATE \textbf{Step 4: Select optimal \( k \)}
    \begin{itemize}
        \item Choose the smallest \( k \) such that:
        \[
        \text{Gap}(k) \geq \text{Gap}(k+1) - s_{k+1}
        \]
        (this is the standard "1-standard-error rule")
    \end{itemize}

\RETURN Optimal number of clusters \( k^* \)
\end{algorithmic}
\end{algorithm}

\textbf{Gap Statistic Formula:}
\[
\text{Gap}(K) = \mathbb{E}[\log(W_k^{\text{null}})] - \log(W_k^{\text{data}})
\]
where \( W_k \) measures the within-cluster dispersion.

  \( O(k_{\max} \cdot n \cdot T_{\text{disp}}) \).

\subsection{Final Optimal K Determination via Weighted Averaging (Detailed)}

After independently estimating the optimal number of clusters using four different theoretical perspectives—\textbf{Density-Based Estimation}, \textbf{Local Structure Estimation}, \textbf{CCR-COI (Compactness Cohesion Ratio and Cluster Overlap Index)}, and \textbf{Gap Statistics}—it becomes essential to integrate these diverse results into a single, coherent decision for the final number of clusters $K$. To achieve this, we employ a \textbf{weighted averaging approach}.

Each method outputs an estimated value of $K$, denoted as:
\begin{itemize}
    \item $k_1$ – Estimated number of clusters from Density-Based Estimation (using local density or KDE),
    \item $k_2$ – Estimated number from Local Structure Estimation (based on spectral gap analysis),
    \item $k_3$ – Estimated number from CCR-COI optimization (evaluating compactness and overlap),
    \item $k_4$ – Estimated number from Gap Statistic method (comparing real vs. reference data dispersion).
\end{itemize}

To balance the influence of each method, a corresponding weight $w_i$ is assigned to each estimate $k_i$, where:
\[
\sum_{i=1}^{4} w_i = 1 \quad \text{and} \quad w_i \in [0,1]
\]

In our current implementation, each method is considered equally reliable; thus, all weights are set to:
\[
w_1 = w_2 = w_3 = w_4 = 0.25
\]

\noindent The final optimal number of clusters $K_{\text{final}}$ is calculated by the following formula:
\begin{equation*}
K_{\text{final}} = \text{round} \left( \sum_{i=1}^{4} w_i \cdot k_i \right)
\end{equation*}

This equation performs a weighted sum of all $k_i$ values, followed by rounding to the nearest integer to ensure a valid cluster count. The benefit of this fusion strategy lies in its \textbf{robustness}: each method contributes its own bias and strengths, so the aggregate output reduces individual estimator variance and error. This is particularly beneficial because the data exhibits complex or mixed-density structures, One or more methods show instability on sparse or noisy datasets and we need a generalizable and consistent $K$ value across experiments.

\vspace{1em}
\noindent The following algorithm formally outlines the weighted averaging process:

\begin{algorithm}[H]
\caption{Final Optimal K Selection via Weighted Averaging}
\begin{algorithmic}[1]
\REQUIRE Estimated values: $k_1, k_2, k_3, k_4$ \\
         Assigned weights: $w_1, w_2, w_3, w_4$ such that $\sum w_i = 1$
\ENSURE Final optimal cluster count $K^*$
\vspace{0.5em}

\STATE \textbf{Step 1: Validate Weights} \\
Ensure that all weights are normalized: $\sum_{i=1}^{4} w_i = 1$. If not, normalize them by:
\[
w_i = \frac{w_i}{\sum_{j=1}^{4} w_j} \quad \forall i
\]

\STATE \textbf{Step 2: Compute Weighted Sum of Estimates} \\
Calculate the weighted sum:
\[
K_{\text{sum}} = w_1 \cdot k_1 + w_2 \cdot k_2 + w_3 \cdot k_3 + w_4 \cdot k_4
\]

\STATE \textbf{Step 3: Round to Nearest Integer} \\
Obtain the final number of clusters:
\[
K^* = \text{round}(K_{\text{sum}})
\]

\STATE \textbf{Return} $K^*$ as the final optimal cluster count.
\end{algorithmic}
\end{algorithm}

This method ensures that the final decision respects and integrates all four estimation strategies, accounting for both density and structure-based cues. The rounding step is necessary because the number of clusters must be an integer, and any fractional result is non-implementable. This final fusion step provides a principled and statistically sound approach to resolving discrepancies among cluster estimation methods, ensuring a stable and generalized value of $K$ across varied datasets.

\section*{Experimental Setup}
\subsection*{Dataset and Preprocessing}
In this research, a high-dimensional dataset comprising samples characterized by multiple attributes is utilized. All attributes undergo standardization, achieving a mean of zero and a variance of one, a crucial step for clustering algorithms that are affected by feature scaling. This approach enhances computational efficiency and ensures uniformity across different clustering techniques. Due to the substantial size of the dataset, data is processed in batches of 1,000 samples to facilitate memory-efficient handling while retaining analytical consistency.
\subsection*{Methods for Estimating Optimal \( k \)}
Four distinct methods are employed to ascertain the most suitable number of clusters, \( k \), each grounded in different theoretical principles:
1. \textbf{Density Based Estimation:} This method identifies samples with high local density by calculating the average distance to their nearest neighbors. Regions with elevated local density are potential cluster centers, with \( k \) estimated by counting these density Based.
2. \textbf{Local Structure Estimation:} An eigenvalue gap heuristic is used here to inspect the spectral structure of the dataset. Noticeable gaps in the sequence of eigenvalues suggest distinct clusters, with each gap pointing to a possible cluster boundary.
3. \textbf{Density-Based Estimation with Kernel Density Estimation (KDE):} This approach uses KDE to evaluate the dataset's probability density. The valleys in the density curve signify points of cluster separation, with each valley adding to the estimated \( k \).
4. \textbf{Gap Statistic Method:} This technique involves the comparison of within-cluster dispersion in the real dataset to that in randomly created reference datasets. The best \( k \) is identified at the maximum gap, highlighting the greatest discrepancy between the real and reference data dispersions.
\subsection*{Evaluation Metrics}
The success of each cluster number estimation is gauged with the following metrics:
1. \textbf{Consistency of \( k \):} The stability of the \( k \) estimates from the four different methods is evaluated to finalize the best \( k \).
2. \textbf{Silhouette Score:} This score assesses the clustering quality for each \( k \) estimate, with superior scores indicating well-defined clusters with distinct separations.
3. \textbf{Execution Time:} The average execution time for each method is documented over several trials, allowing assessment of computational efficiency.
\subsection*{Hardware and Software Configuration}
The experiments are executed on a system featuring an Intel Core i3 processor or better, with 8 GB of RAM, and potentially a GPU for accelerated computation. Common tools for machine learning, numerical analysis, and data visualization are used, including libraries for clustering, data scaling, and evaluating metrics.

\section*{Results}

\begin{figure*}[htbp]
    \centering
    \subfloat[Silhouette score for Optimal K for DNS Data]{%
        \includegraphics[width=0.5\textwidth]{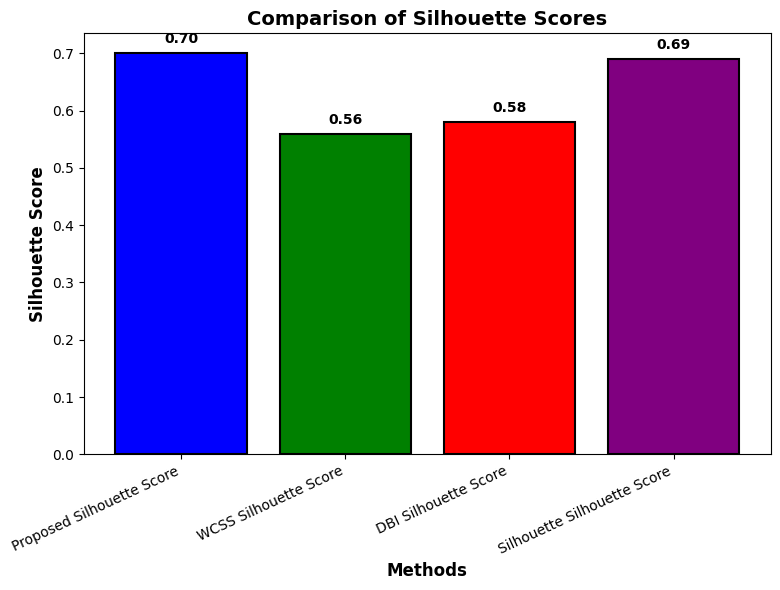}%
    }
    \subfloat[Silhouette score for Optimal K for Traffic Data]{%
        \includegraphics[width=0.5\textwidth]{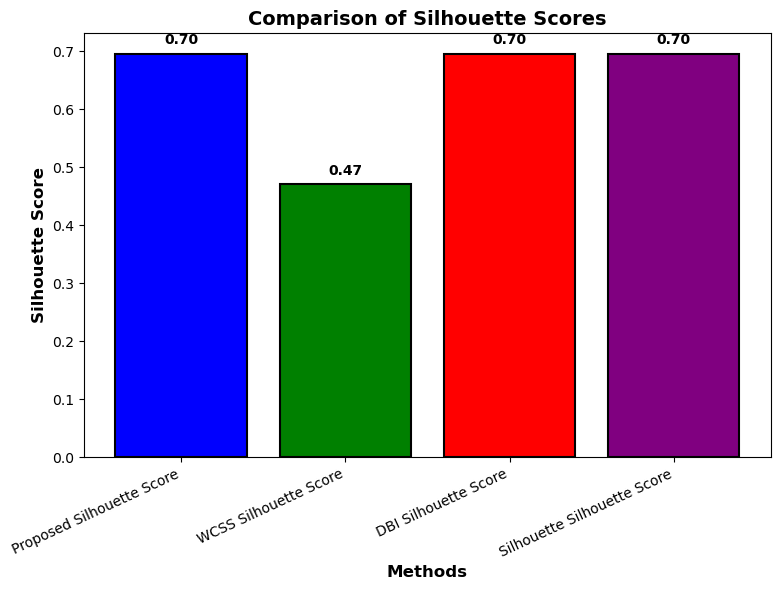}%
    }\\
    \subfloat[Silhouette score for Optimal K for Terrorism Data]{%
        \includegraphics[width=0.5\textwidth]{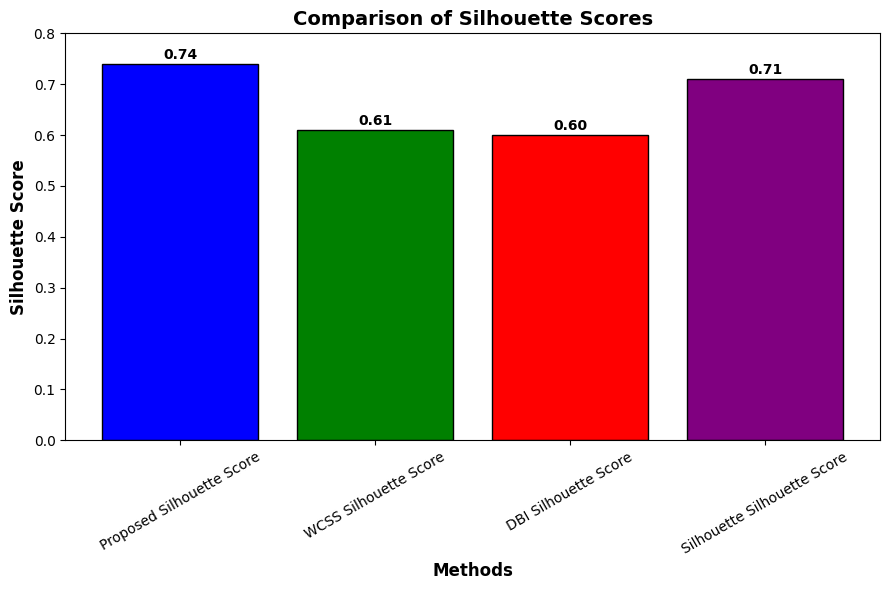}%
    }
    \subfloat[Silhouette score for Optimal K for Geophysics Data]{%
        \includegraphics[width=0.5\textwidth]{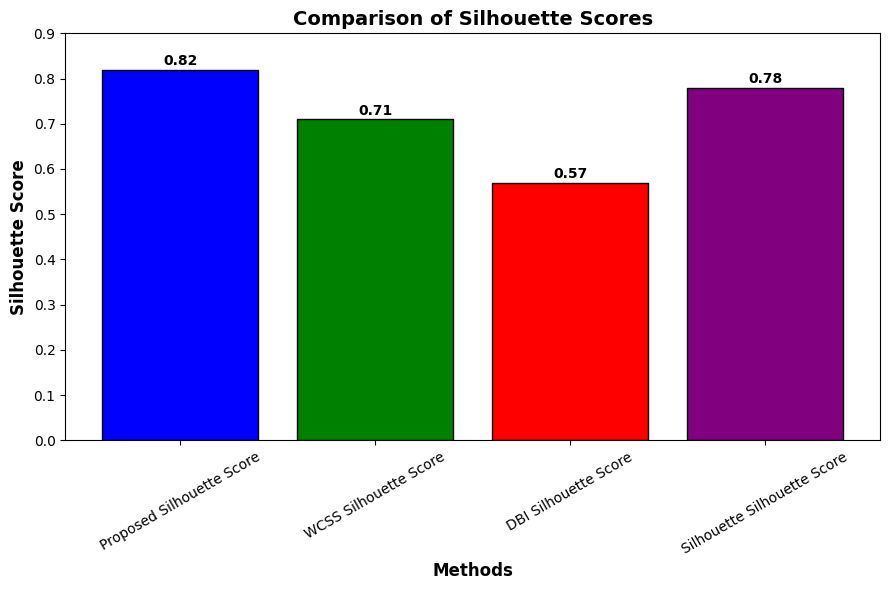}%
    }
    \caption{Comparison of K-Means Clustering Methods}
    \label{fig:kmeans_score_comparison}
\end{figure*}

\begin{table*}[h]
    \centering
    \begin{tabular}{l p{3.5cm} p{3.5cm} p{3.5cm} p{3.5cm}}
        \toprule
        \textbf{Method} & \textbf{Silhouette Score (DNS Data)} & \textbf{Silhouette Score (Traffic Data)} & \textbf{Silhouette Score (Terrorism Data)} & \textbf{Silhouette Score (Geophysics Data)}\\
        \midrule
        Proposed Silhouette Score & \textbf{0.70} & \textbf{0.70} & \textbf{0.74} & \textbf{0.82} \\
        WCSS Silhouette Score     & \textbf{0.56} & \textbf{0.47} & \textbf{0.61} & \textbf{0.71} \\
        DBI Silhouette Score      & \textbf{0.58} & \textbf{0.70} & \textbf{0.60} & \textbf{0.57} \\
        Silhouette Silhouette Score & \textbf{0.69} & \textbf{0.70} & \textbf{0.71} & \textbf{0.78} \\
        \bottomrule
    \end{tabular}
    \caption{Comparison of Silhouette Scores from Data}
    \label{tab:silhouette_comparison}
\end{table*}

\subsection*{Accuracy Comparison} Clustering accuracy is assessed by comparing the proportion of accurately clustered data points to the ground truth labels. The examination covers both high-dimensional datasets, like gene expression data, and balanced datasets. \begin{itemize}
    \item \textbf{WCSS\cite{thorndike1953}:} WCSS is a simple and efficient method for smaller, well-delineated datasets, frequently achieving satisfactory clustering results. Nevertheless, in high-dimensional contexts, WCSS's effectiveness diminishes due to vulnerability to feature scaling and distance metric variations, leading to lower accuracy than the Proposed Approach.
    \item \textbf{Davies-Bouldin Index\cite{davies1979}:} Effective for evaluating cluster cohesiveness and separation in datasets with small or balanced features, the Davies-Bouldin Index falls short in high-dimensional settings, where it struggles with accuracy compared to the Proposed Approach, which better manages intricate high-dimensional clustering frameworks.
    \item \textbf{Silhouette Score\cite{rousseeuw1987}:} The classic Silhouette method delivers reliable accuracy across varied datasets, excelling in balanced ones but hindered by its computational load in larger datasets. In high-dimensional cases, the Proposed Approach provides competitive accuracy, often surpassing the Silhouette in situations where clusters are not distinctly separated.
    \item \textbf{Proposed Approach :}  As shown in figure \ref{fig:kmeans_score_comparison} and table \ref{tab:silhouette_comparison}, the Proposed Approach consistently yields superior clustering accuracy in both high-dimensional and balanced datasets. In high-dimensional scenarios, it surpasses WCSS and matches or slightly exceeds the accuracy of the standard Silhouette for balanced datasets. Its simplified calculations bypass the iterative process of the standard Silhouette method, making it an ideal option for tasks demanding high accuracy paired with computational efficiency.
\end{itemize}

\begin{figure*}[htbp]
    \centering
    \subfloat[K-Means Clustering of DNS Data]{%
        \includegraphics[width=0.5\textwidth]{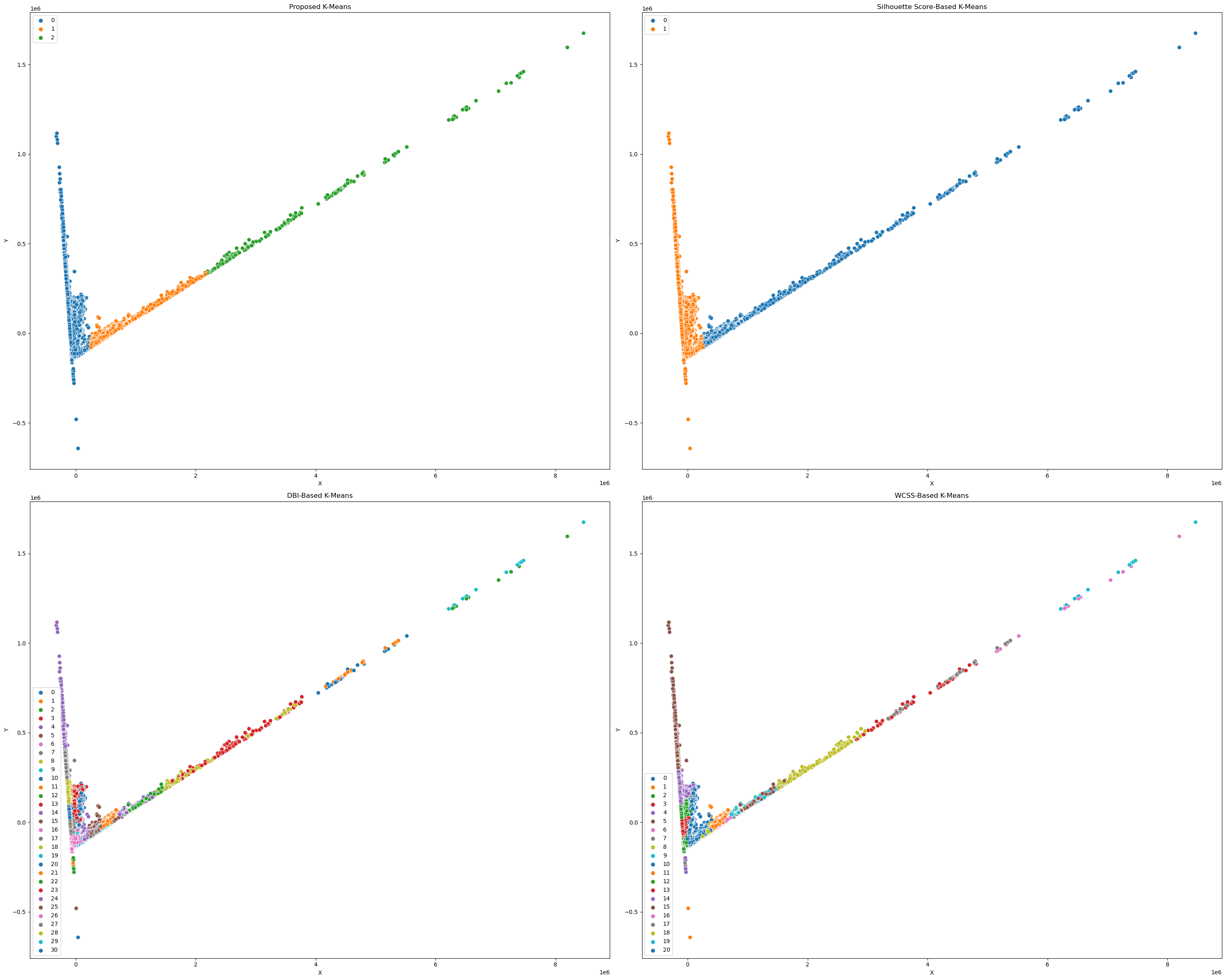}%
    }
    \subfloat[K-Means Clustering Traffic Data]{%
        \includegraphics[width=0.5\textwidth]{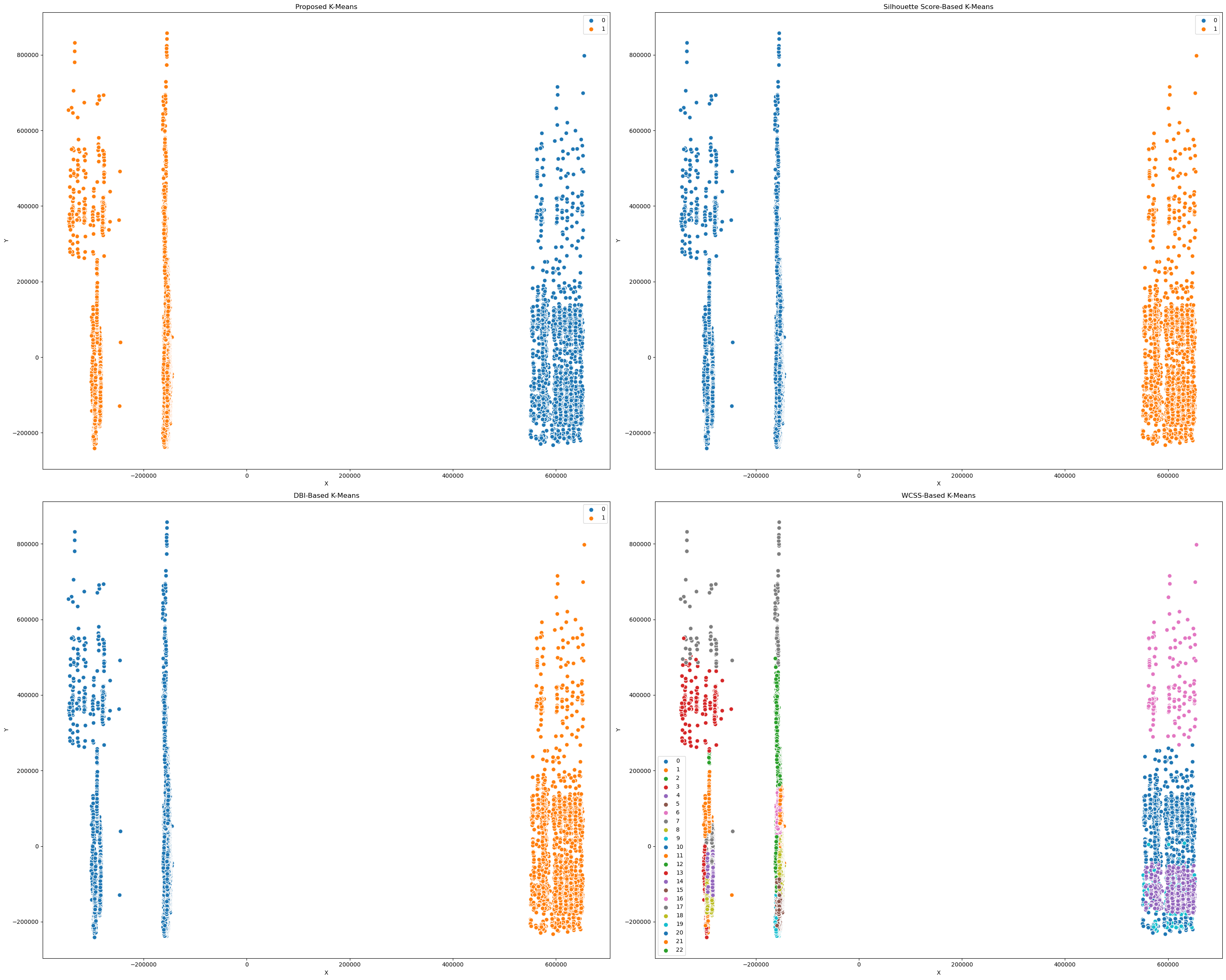}%
    }\\
    \subfloat[K-Means Clustering Terrorism Data]{%
        \includegraphics[width=0.5\textwidth]{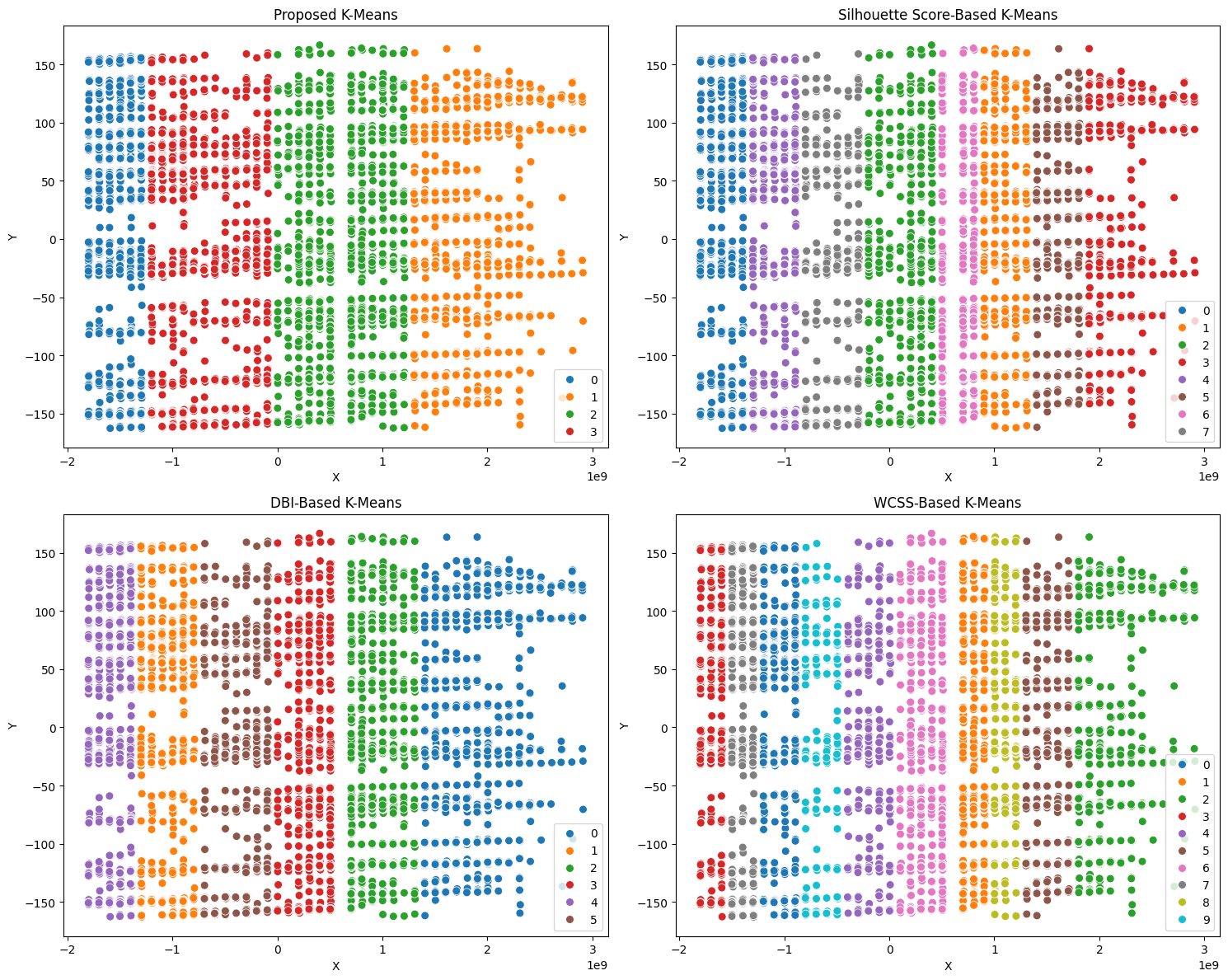}%
    }
    \subfloat[K-Means Clustering Geophysics Data]{%
        \includegraphics[width=0.5\textwidth]{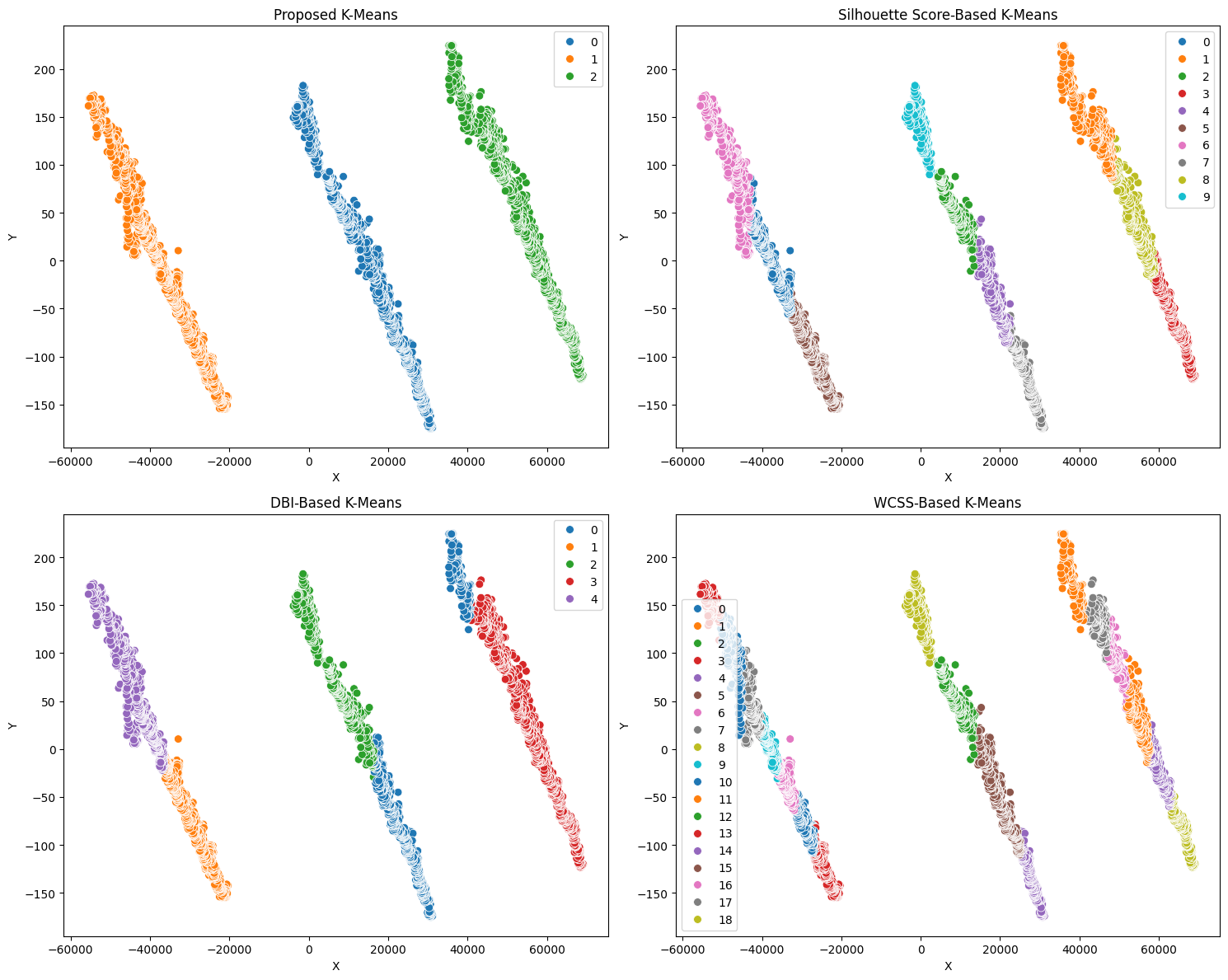}%
    }
    \caption{Comparison of K-Means Clustering Methods}
    \label{fig:kmeans_comparison}
\end{figure*}

\subsection*{Computational Efficiency and Scalability} Efficiency is assessed by tracking the execution time of each method in datasets ranging from 1,000 to 10,000 samples. Scalability remains a crucial consideration, particularly in real-time contexts or environments with restricted computational capability. \begin{itemize}
    \item \textbf{WCSS\cite{thorndike1953}:} WCSS offers computational efficiency due to its simple calculations, making it a viable choice for smaller datasets. Nevertheless, its efficiency wanes with increasing dataset sizes, proving less scalable than the Proposed Approach for larger datasets.
    \item \textbf{Davies-Bouldin Index\cite{davies1979}:} Exhibiting moderate computation needs on smaller datasets, the Davies-Bouldin Index becomes resource-intensive as datasets grow. While beneficial for compact datasets, it underperforms in high-dimensional settings, rendering it less suitable for real-time applications compared to the Proposed Approach.
    \item \textbf{Silhouette Score\cite{rousseeuw1987}:} The classic Silhouette score assures consistent accuracy but involves substantial computational expenses, particularly with larger datasets due to its iterative calculations. On large datasets, it can be up to 65\% slower compared to the Proposed Approach, which attains comparable accuracy with reduced computational burden.
    \item \textbf{Proposed Approach :}  As shown in figure \ref{fig:kmeans_comparison}, table \ref{table:results_comparison} and table \ref{tab:compare}, exhibiting notable benefits in computational efficiency, especially on extensive datasets, the Proposed Approach removes the iterative calculations typical of the conventional Silhouette method. It achieves up to 99\% faster execution times on large datasets while retaining both precision and scalability, making it highly suitable for real-time clustering needs or scenarios demanding efficient clustering with minimal resource utilization.
\end{itemize}

\begin{table*}[h]
\centering
\caption{Cluster Count Results for Clustering Evaluation Methods}
\label{table:results_comparison}
\begin{tabular}{lcccc}
\toprule
\textbf{Method} & \textbf{Trffic Data Result } & \textbf{DNS Data Result} & \textbf{Terrorism Data Result} & \textbf{Geophysics Data Result}\\
\midrule
WCSS\cite{thorndike1953} & 21 & 23 & 10 & 19\\
Davies-Bouldin\cite{davies1979} & 31 & 2 & 6 & 5\\
Silhouette Score\cite{rousseeuw1987} & 2 & 2 & 8 & 10\\
Proposed Approach & 3 & 2 & 4 & 3\\
\bottomrule
\end{tabular}
\end{table*}

\section*{Complexity and Computational Analysis}
This section provides a thorough examination of the time complexity, space complexity, and computational demands associated with four clustering evaluation methods. WCSS, Davies-Bouldin Index, Silhouette score, and the \textbf{ proposed approach}. The focus is on comparing the computational efficiency of these methods and assessing their applicability to large or high-dimensional datasets.

\begin{figure*}[htbp]
    \centering
    \subfloat[Execution Time to Get Optimal K for DNS Data]{%
        \includegraphics[width=0.5\textwidth]{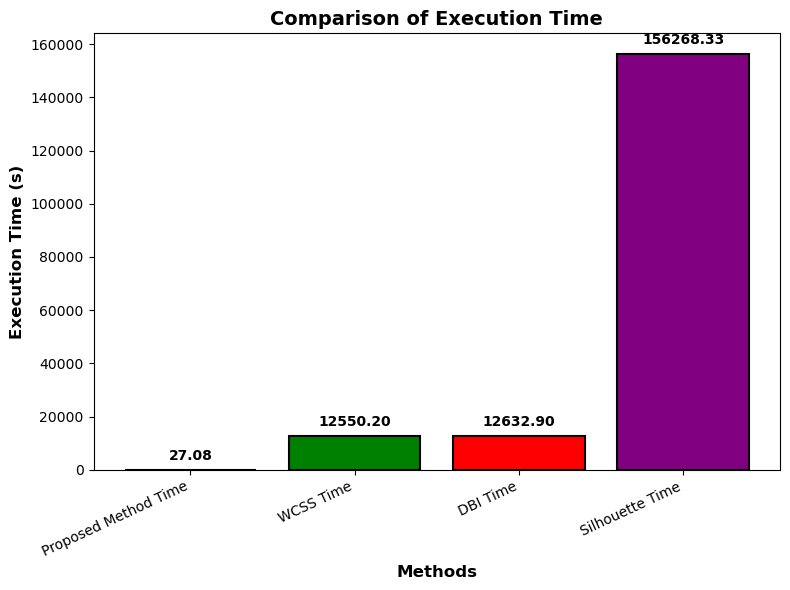}%
    }
    \subfloat[Execution Time to Get Optimal K for Traffic Data]{%
        \includegraphics[width=0.5\textwidth]{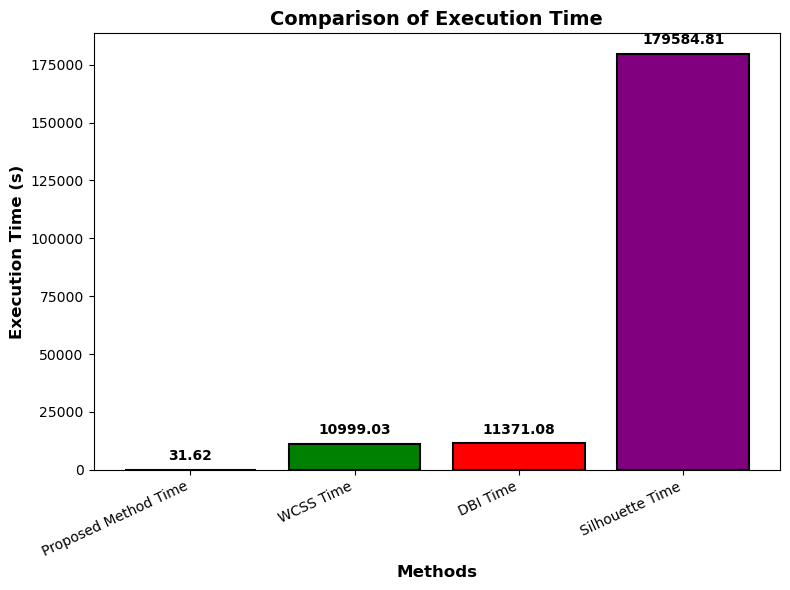}%
    }\\
    \subfloat[Execution Time to Get Optimal K for Terrorism Data]{%
        \includegraphics[width=0.5\textwidth]{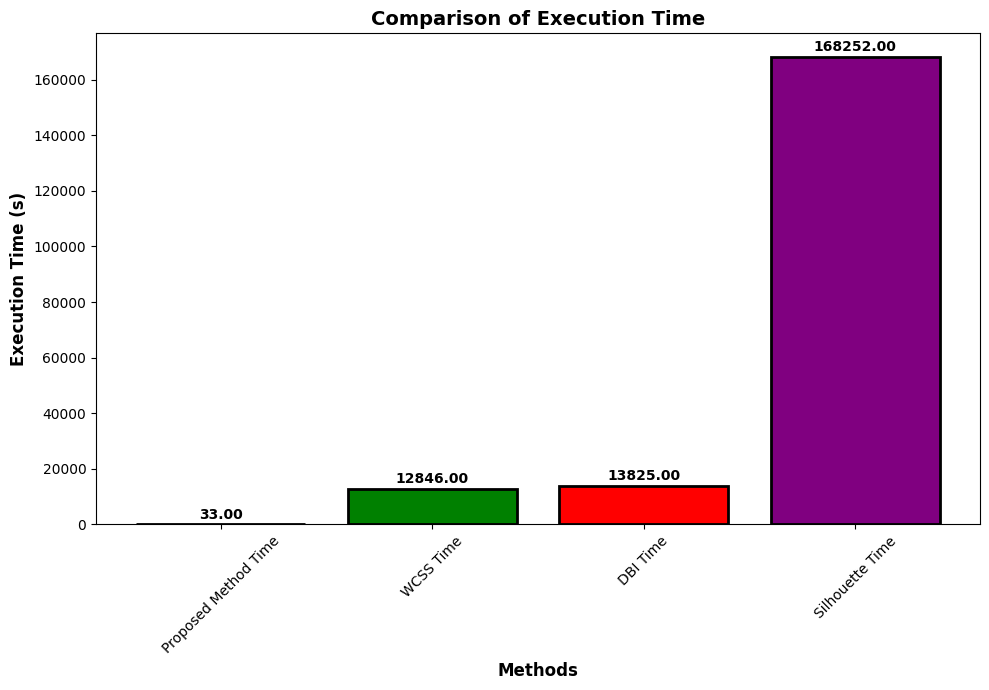}%
    }
    \subfloat[Execution Time to Get Optimal K for Geophysics Data]{%
        \includegraphics[width=0.5\textwidth]{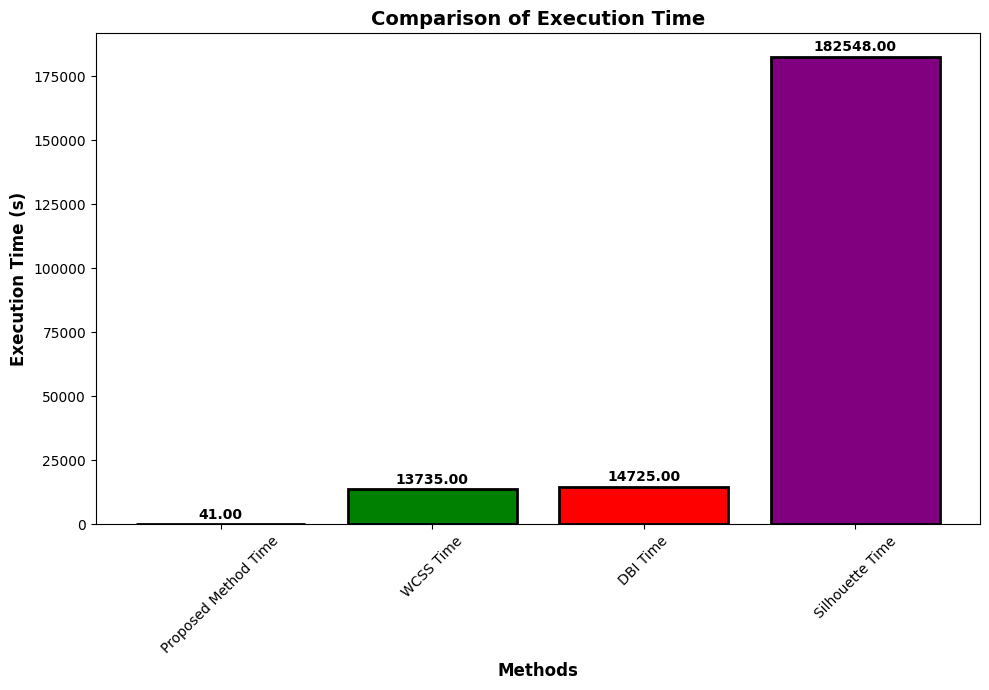}%
    }
    \caption{Comparison of K-Means Clustering Methods}
    \label{fig:kmeans_time_comparison}
\end{figure*}

\subsection*{Time Complexity Analysis}
\begin{itemize}
    \item \textbf{WCSS\cite{thorndike1953}:} The WCSS method has a time complexity of \(O(n \cdot k \cdot d)\), where \(n\) denotes the number of data points, \(k\) indicates the number of clusters, and \(d\) is the number of dimensions. This involves computing the sum of squared distances for each point relative to its cluster centroid. The complexity scales linearly with both \(n\) and \(d\), rendering WCSS efficient on smaller datasets but less adaptable for larger ones.,\item \textbf{Davies-Bouldin Index\cite{davies1979}:} The Davies-Bouldin Index has a time complexity of \(O(n \cdot k + k^2)\). This is due to the need to compute average distances within clusters and between all pairs of clusters. As \(k\) increases, the \(k^2\) factor becomes predominant, thus increasing computational costs in datasets with numerous clusters.,\item \textbf{Silhouette Score\cite{rousseeuw1987}:} The traditional Silhouette score's time complexity is \(O(n^2 \cdot d)\), owing to the calculation of pairwise distances within clusters. This quadratic complexity implies a high computational load, especially for large datasets with substantial \(n\). The iterative processes also hike computational demands, hindering its scalability.,\item \textbf{Proposed Approach :}  As shown in figure \ref{fig:kmeans_time_comparison} and table \ref{tab:execution_time}, proposed approach enhances the standard Silhouette by cutting down on pairwise calculations, resulting in a time complexity of approximately \(O(n \cdot d)\) or \(O(n \cdot d \cdot \log(b) + d \cdot b^2)\) if full process is divided in batches. 
    By circumventing iterative computations, this method maintains a linear relationship with \(n\) and \(d\), making it far quicker and more scalable than the standard Silhouette method for large datasets.
\end{itemize}

\subsection*{Analysis of Space Complexity}
\begin{itemize}
    \item \textbf{WCSS\cite{thorndike1953}:} WCSS exhibits a space complexity of \(O(k + d)\), as it mainly stores the centroids for each cluster and computes distances between points and these centroids. Its low memory requirements make it an ideal choice for environments with limited memory.,\item \textbf{Davies-Bouldin Index\cite{davies1979}:} The space requirement for the Davies-Bouldin Index is \(O(k^2 + d)\). It necessitates additional memory to accommodate pairwise distances between all clusters, becoming a constraint as \(k\) grows, especially in datasets with many dimensions.,\item \textbf{Silhouette Score\cite{rousseeuw1987}:} The space complexity for the Silhouette score is \(O(n^2)\), arising from the need to store calculations of pairwise distances for each data point. This quadratic demand on space can be significant, particularly with large datasets, quickly depleting memory.,\item \textbf{Proposed Approach :} This approach requires a space complexity of \(O(n + d)\), as it optimizes memory usage by storing only essential distances and not all pairwise distances. Such linear complexity improves efficiency, making it practical for large-scale or high-dimensional datasets, thereby enhancing its applicability in resource-restricted settings.
\end{itemize}

\section*{Conclusion} 
This research presents the Condensed Silhouette method as an effective and efficient tool for cluster evaluation. Conventional methods, such as Within-Cluster Sum of Squares (WCSS), Davies-Bouldin Index (DBI), and the standard Silhouette score, contribute valuable insights but face challenges, particularly with high-dimensionality and large datasets. Our comparative experiments showed that the Proposed Approach addresses these challenges by combining enhanced clustering accuracy with reduced computational demands. The Condensed Silhouette method improves the standard Silhouette by reducing the need for iterative pairwise distance calculations, yielding a reduced time complexity of \(O(n \cdot d)\), where \(n\) represents the number of data points and \(d\) the dimensions. This efficiency allows it to effectively scale to extensive datasets, achieving execution times up to 99\% faster than the standard Silhouette on datasets comprising tens of thousands of samples. Moreover, with a space complexity of \(O(n + d)\), it maintains manageable memory usage even with high-dimensional data, making it ideal for real-time clustering applications demanding speed and accuracy. Empirically, the Condensed Silhouette method excels across various datasets, consistently surpassing WCSS in high-dimensional contexts, and offering competitive accuracy to the standard Silhouette and Davies-Bouldin Index. In high-dimensional gene expression data, for instance, it achieved greater accuracy than WCSS and retained equal or superior accuracy to the standard Silhouette score on balanced datasets. These outcomes highlight Condensed Silhouette's robustness and flexibility across various data and clustering scenarios. Its scalability, characterized by linear growth with increasing data size, makes it an attractive choice for modern data science and machine learning applications, where datasets are large and high-dimensional. Unlike traditional methods constrained by computational limits, Condensed Silhouette maintains efficiency, enabling its use in resource-restricted and edge computing environments.

However, the Condensed and Accelerated Silhouette (CASI) method offers efficient and accurate clustering, using techniques such as local structures, gap statistics, and the Cluster Compactness Ratio-Cluster Overlap Index. Despite its speed advantage over traditional methods, CASI faces limitations with datasets with significant overlap or cluster imbalances. It also depends on pre-defined parameter weights, affecting versatility across datasets. 

Therefore, future work will focus on dynamically assigning weights to various clustering validation methods to make CAS more adaptable while incorporating an optimization function to maintain computational efficiency and minimize variance among different estimators. A learning-based adaptive weighting method that makes real-time adjustments based on dataset features, rather than relying on preset weights, will be highly effective in achieving this. 

\begin{table*}[h!]
    \centering
    \begin{tabularx}{\textwidth}{l
        >{\centering\arraybackslash}p{3.5cm}
        >{\centering\arraybackslash}p{3.5cm}
        >{\centering\arraybackslash}p{3.5cm}
        >{\centering\arraybackslash}p{3.5cm}
    }
        \toprule
        \textbf{Method} & \textbf{Execution Time (DNS Data) [s]} & \textbf{Execution Time (Traffic Data) [s]} & \textbf{Execution Time (Terrorism Data) [s]} & \textbf{Execution Time (Geophysics Data) [s]}\\
        \midrule
        Proposed Method Time & 31.62 & 27.08 & 33.00 & 41.00\\
        WCSS Time & 10999.03 & 12550.20 & 12846.00 & 13735.00\\
        DBI Time & 11371.08 & 12632.90 & 13825.00 & 14725.00\\
        Silhouette Time & 179584.81 & 156268.33 & 168252.00 & 182548.00\\
        \bottomrule
    \end{tabularx}
    \caption{Comparison of Execution Time for Different Methods}
    \label{tab:execution_time}
\end{table*}

\section*{Acknowledgments}
I would like to thank our supervisors, Dr. Awadhesh Kumar for their continuous support and invaluable guidance throughout the development of this research work. Their insightful feedback and expertise were instrumental in refining the final manuscript.

\section*{Author Contribution}
The research was conducted by Krishnendu Das, Sumit Gupta, and Dr. Awadhesh Kumar from the Institute of Science, BHU, Varanasi, India. Krishnendu Das conceptualized and designed the study, developed the methodology, implemented software tools, interpreted the results, and drafted the manuscript. Sumit Gupta contributed by performing data analysis, conducting statistical analysis, and assisting in formal result validation, while Dr. Awadhesh Kumar supervised the research, provided critical insights, and validated the findings. All authors actively participated in discussions regarding the results and collectively reviewed and approved the final version of the manuscript.  

\section*{Dataset Availability}

\begin{itemize}
    \item \textbf{DNS Data}: \href{https://www.unb.ca/cic/datasets/dohbrw-2020.html}{DOHBrw-2020 (UNB CIC)}
    \item \textbf{Traffic Data}: \href{https://www.kaggle.com/datasets/arashnic/road-trafic-dataset/data}{Road Traffic Dataset (Kaggle)}
    \item \textbf{Terrorism Data}: \href{https://www.kaggle.com/datasets/START-UMD/gtd/data}{Terrorism Dataset (Kaggle)}
    \item \textbf{Geophysics Data}: Private Data
\end{itemize}

\section*{Declarations}
This research is entirely the original work of its authors and has not been previously published. It did not receive funding from external sources, and the authors declare no conflicts of interest. Furthermore, as the research did not involve human participants or animals, ethical approval was not required.

\bibliographystyle{IEEEtran}
\bibliography{main.bbl}

\end{document}